**Title:** Robot Tape Manipulation for 3D Printing

**Authors:**
Nahid Tushar,[1] Rencheng Wu,[1] Yu She,[2] Wenchao Zhou,[1] Wan Shou[1]*

**Affiliations:**
[1]Department of Mechanical Engineering, University of Arkansas; Fayetteville, AR 72701, USA.
[2] Department of Industrial Engineering, Purdue University; West Lafayette, IN 47907, USA.

*Corresponding author. Email: wshou@uark.edu.

**Abstract:** 3D printing has enabled various applications using different forms of materials, such as filaments, sheets, and inks. Typically, during 3D printing, feedstocks are transformed into discrete building blocks and placed or deposited in a designated location similar to the manipulation and assembly of discrete objects. However, 3D printing of continuous and flexible tape (with the geometry between filaments and sheets) without breaking or transformation remains underexplored and challenging. Here, we report the design and implementation of a customized end-effector, i.e., tape print module (TPM), to realize robot tape manipulation for 3D printing by leveraging the tension formed on the tape between two endpoints. We showcase the feasibility of manufacturing representative 2D and 3D structures while utilizing conductive copper tape for various electronic applications, such as circuits and sensors. We believe this manipulation strategy could unlock the potential of other tape materials for manufacturing, including packaging tape and carbon fiber prepreg tape, and inspire new mechanisms for robot manipulation, 3D printing, and packaging.

**One-Sentence Summary:** Robot-assisted manipulation of tape materials for 3D printing.

**Main Text:**

**INTRODUCTION**

Additive manufacturing (AM, a.k.a. 3D printing) has advanced product innovation in the past decades, benefiting both structure design and function integration (*1, 2*). Progress has been made to innovate printing materials and printing processes, in terms of building blocks, joining mechanisms, forms of control, and transformation methods. Typically, material forms for 3D printing include solid filaments, wires, liquid resins, powders, and sheets (*1*). These feedstocks are transformed into discrete building units (such as droplets and lines) and placed, deposited, or solidified at designated locations for layer-by-layer manufacturing. However, 3D printing of continuous and flexible tape (with the geometric form in between filaments and sheets) without breaking or transformation remains underexplored and challenging. In the composite manufacturing industry, carbon fiber prepreg tapes are widely used for placement, which is called automated tape placement/laying (ATP/ATL) (*3*). Such ATP systems generally use heat and pressure to consolidate the composite materials (*4, 5*). However, ATP/ATL systems are typically mounted with large-scale gantry systems or robotic arms (*4, 6–8*). Such approaches require high capital investment and complex heavy equipment, which is not easily accessible to general researchers and difficult to integrate with desktop-scale 3D printing technologies. Due to the heavy and complex tooling (*3, 9, 10*), conformal placement of tape without wrinkle is challenging on curved surfaces (*11–13*). Also, the placement of tape is not suitable for small-scale object fabrication. Furthermore, out-of-plane 3D printing of tape materials has not been realized before (*3*). Lastly, ATP systems typically work with carbon fabric/fiber prepreg tapes almost exclusively without extending to other tape materials. Therefore, enabling AM of various tape materials can diversify existing manufacturing capabilities, and advance the intersection of manufacturing and robotics, thereby promoting Industry 4.0 (*3, 14, 15*).

So far, conventional tapes have seen limited usage in additive manufacturing (AM). Due to their low cost, and unique geometry, different tapes are used for packaging (*16–18*), repairing (*19*), and flexible electronics (*20–24*), etc. However, there is no existing automated approach to place and transform tapes into more complex geometries and functionalities. A few challenges can be envisioned with this new form of printing material: (a) tension inside the continuous tape material should be carefully managed; (b) the feeding and placement apparatus should be reasonably small for desktop-scale fabrication; (c) precise cutting mechanism for tape is required; and (d) manipulation and placement should be generalizable for different tape materials. One promising and commonly used tape material is copper (Cu) tape, which has excellent conductivity. Although copper is widely utilized as conductors, there is no research available on using Cu tape as feedstock for 3D printing.

On the other hand, robotic hands are being extensively researched to achieve more accurate and intelligent manipulation (*25, 26*); however, there are limited manipulators and reports on tape manipulation from the robotics perspective. Deformable linear objects (DLOs) are of great interest and importance to robotic manipulation, and different manipulations have been demonstrated, such as untangling (*27*), and dynamic manipulation (*30*). During the manipulation, the friction force is considered as a critical parameter to maintaining robust grasping of DLOs (*31*). Zhu, et al. (*32*) also harnessed the contacts for cable shaping, which is inspiring for material structuring. Yet, all of these reports focus on the dexterous manipulation of DLOs and none of them consider using a manipulation strategy for additive manufacturing with tape.

Recently, a few papers reported deformable object manipulation for advanced manufacturing. For example, Kayser, et al. reported a fiber robot, where fiber was interwoven to make composite tubes for complex forms through automated winding on an inflatable silicone tube (as the support) (*33*). Augugliaro, et al. reported (*34*) building tensile structures with quadcopters, where ropes, cables, and wires were used as the construction elements. They noticed that tension forces played an important role during the construction. From the perspective of manipulator, Jiang et al. reported a roller-based gripper for cable harvesting (*35*). Yuan, et al. reported a roller-based grasper (*36*), where motorized rollers were used to realize grasping and manipulation. This work inspires us to consider different types of manipulators (*35, 36*) and re-think ATP from the perspective of robotic manipulation.

To realize tape 3D printing, it is necessary to consider new mechanisms and draw insights from robot manipulation and material characteristics. In this work, for the first time, we report a manipulation strategy for tape placement and 3D printing. To realize out-of-plane tape placement and 3D printing, we first designed, and fabricated a custom tape print module (TPM) and mounted it to a 6-axis robot (UR10e from Universal Robotics) as an end effector, as shown in Fig. 1 A and B (details can be found in Materials and Methods, and note S1). The system control architecture is illustrated in Fig. 1C, which is composed of toolpath generation, robot controller, and print control module (PCM). The custom TPM enables feeding, compaction, and cutting of the tape (detailed schematics can be found in fig. S1) while the robot follows a defined toolpath for tape placement and 3D printing. With this system, we realized the printing of various tape materials on different substrates to generate different complex geometric shapes in and out of the plane. We further demonstrated that Cu tape functions as a conductor for high-current electrical components and out-of-plane circuits. By using multiple tape materials, we fabricated a capacitive sensor array for human-robot interactions. 3D printing capability was also demonstrated by layer-by-layer placement of tape to fabricate a woodpile structure. Our system combines robot tape manipulation for additive manufacturing, which represents a unique desktop-scale 3D printing approach using dexterous robots.

**Fig. 1. Overview of tape manipulation and 3D printing system.** (**A**) Assembled tape placement module (TPM) on a 6-axis robot. (**B**) Photograph of the TPM end-effector. (**C**) System architecture overview. (More details can be found in fig. S2).

# RESULTS
## Planar placement
To demonstrate the planar placement and printing capabilities of our proposed system, we researched various common geometries. Figure 2A demonstrates copper tape of different widths can be readily placed on substrates by adjusting the rollers of the TPM. Meanwhile, various other tape materials, such as vinyl tape, and polyimide (PI) tape, are printed by switching the feedstock. In addition to simple straight lines, the dexterous robotic arm, equipped with six degrees of freedom (DOF), enables the printing of wavy lines with varying amplitudes and wavelengths following a nonlinear toolpath (Fig. 2B, and movie S1). We demonstrate printing complex geometries using two methods: printing line-by-line segments (Fig. 2C(i-ii)) and continuous printing (Fig. 2C (iii-v)). It is observed that due to the rigidity of Cu tape across its width, stress can be induced while printing on a nonlinear toolpath causing wrinkles on the tape surface (fig. S3). However, this issue can be mitigated by updating the toolpath with an increased curvature radius. It is worth noting that circular trajectory placement is challenging for both robots and humans (*37, 38*). Nevertheless, our TPM can accomplish relatively uniform tape placement on circular toolpaths repeatedly. Figure 2C(iii) shows a set of printed concentric circles with different diameters. It is found that during the printing of circles, with the decrease in diameter, the printing becomes more challenging, typically causing more wrinkles. Besides, printing complete circle traces are constrained by the robot joint limits which affect the start and end points quality (as shown in fig. S4, the smallest diameter successfully printed with our system is 5 cm). As the adhesive on the copper tape is strong enough, the printing can be conducted on different substrates, as shown in Fig. 2D with letters of "UARK" on PI, wood, acrylic, and metal, regardless of their surface roughness and chemical properties.

**Fig. 2. Representative Cu tape 2D printing.** (**A**) Straight lines of Cu tapes (with different widths), and other tape materials (e.g., polyimide and vinyl). (**B**) Nonlinear printing of wavy lines. (**C**) Common geometries printed with Cu tape: segment-by-segment printing of a (i) triangle and (ii) rectangle; continuous printing of (iii) circles, (iv) hexagon, and (v) rectangle; (vi) a smiley face. (**D**) Cu tape printed "UARK" on different substrates. (scale bar, 3 cm.)

## System study: influence of processing parameters

In our study, the tape is treated as a deformable object that is manipulated between a contact substrate (or point) and the compaction roller. In order to gain a better understanding of the manipulation, several key processing parameters are studied, including the relative motion speed between roller and substrate (namely, print speed), compaction force, and adhesion force between Cu tape and substrate. We first evaluate the reliability of manipulation by printing different lengths (from 5 to 20 cm). Figure 3A-B demonstrates that there is a reasonable agreement between the designed and printed copper tape on three common substrates (acrylic, metal, and wood) regarding total length and straightness (further measurement details can be found in note S2, and fig. S5). It is also observed that with the increase of length from 5 to 20 cm, the mean length becomes closer to the design (Fig. 3A inset). However, no clear difference in length deviation is observed due to the different substrates. Due to the ductility and deformability of Cu tape, the cutting blade of the TPM often leads to uneven or irregular cuts, which contributes to the length and straightness deviation. Minor straightness deviation (with a ratio to printed length of ~0.2%) is noticed during printing (Fig. 3B), which accumulates with the increase of printed length. A slight influence of the substrate on print straightness is noticed which could be associated with adhesion between the tape and different substrates. The TPM's rigidity along with the acceleration and deceleration of the robotic arm's motion, and robot position repeatability play critical roles in dimensional accuracy (*39–41*).

We also tested different printing speeds that are comparable to fused filament fabrication (FFF) printers (*42*), ranging from 10 mm/s to 100 mm/s, and a reasonable printing accuracy was achieved although the corresponding values increased slightly (~3% in total length and ~6% tape width) with the increase in speed (Fig. 3, C and D). It should be noted that high print speed and high compaction force may cause system instability and require frequent system calibration (such as tension, alignment, z-height, etc.) before printing. Regarding toolpath generation, we conducted a study comparing Cartesian point-to-point print path planning (without control of compaction force on the substrate) to relative point-to-point path planning (with controlled compaction force applied normally to the print substrate) (additional information can be found in note S3). However, no significant advantages or disadvantages were observed between these two methods in terms of dimensional accuracy or the surface roughness of the printed tape (additional information on note 3, and fig. S6). Furthermore, we explored the impact of compaction force magnitude, ranging from 2 to 8 N, with a 15 cm copper tape placement task (as shown in Fig. 3E), the straightness deviation of the placed tape initially decreases and then increases with increasing compaction force, with an optimal value of 4 N. No clear influence was observed regarding compaction force on print length for planar placement (fig. S7).

Additionally, the printed tape's surface roughness and adhesion on different print substrates were examined. Print samples with 15 cm lengths were printed on acrylic, metal, and wood substrates for analysis (details of measurement can be found in Materials and Methods, and fig. S8). Figure 3F suggests a direct correlation between the surface smoothness of printed copper tape and the surface roughness of the print substrate. Wood substrate shows the highest roughness, while metal shows the lowest roughness, and the surface smoothness of printed tape on these substrates follows the same trend. Correspondingly, the adhesion force between the tape and the substrate inversely correlates with the surface roughness of the substrates. However, the influence of different substrates on printing quality remains inconclusive. Further studies with a variety of substrates could be conducted to better understand their effects on print quality. We believe the achieved accuracy in this study is suitable for potential applications, such as tape placement in packaging (*43*).

**Fig. 3. The influence of Cu tape printing parameters on quality.** (**A**) print length consistency (print speed 25 mm/s, point-to-point toolpath trajectory); (**B**) straightness consistency (print speed 25 mm/s, point-to-point toolpath trajectory); (**C**) print length consistency and (**D**) straightness consistency with different print speeds with a fixed length of 15 cm (point-to-point toolpath trajectory); (**E**) print straightness consistency with variable compaction force on the substrate (relative point to point toolpath trajectory with compaction force applied normal to the substrate); (**F**) tape adhesion and roughness on different substrates (details of measurement can be found in Materials and Methods section and fig S8).

**Non-planar placement and 3D printing**
Further capabilities of our tape placement system were explored and demonstrated beyond planar tape placements. By leveraging the tension between the substrate and the TPM contacts (*32*), tape placement is achieved on non-planar or multi-planar surfaces, as shown in Fig. 4A, where overhanging structures are fabricated without additional support. Due to the flexibility of the robotic arm (with 6 DoFs), tape placement on a vertical surface (or wall) can be readily achieved (Fig. 4B, movie S2). Additionally, employing proper robot manipulation, our system achieved conformal placement of copper tape on a hemisphere surface without tape breakage or surface wrinkles (Fig. 4C, movie S3). Due to the unique continuous nature of Cu tape and robot dexterity, electrical conductors can be readily printed on non-planar surfaces (such as vertical surfaces and out-of-plane objects), without the need for creating channels, holes, or traces on a part (*44, 45*) or post-processing (*46, 47*) using conductive inks or filaments. Conventional conductive filaments and solution-based inks often require post-processing to achieve optimal properties and overhang structures or vertical surface printing is not achievable with such feedstocks (*48*). Additionally, with advanced toolpath planning and manipulation of tension between two contact points of the tape, three-dimensional structures (such as a simple 3D wall, an overhanging stack, and a woodpile structure, Fig. 4D) are demonstrated through layer-by-layer placement of the copper tape. A microscope image (Fig. 4D (iii)) reveals that the tape has a dense structure and layer feature, similar to many other 3D printed structures. It is important to note that conformal printing (in Fig. 4C) and layer-by-layer 3D printing were achieved through active compaction force control exerted normally to the substrate contact surface with the TPM. While earlier sections revealed no apparent advantages or disadvantages of applying active compaction force control in various tape printing scenarios, it becomes crucial for conformal printing on curved surfaces and 3D structure printing.

**Fig. 4. Printing beyond traditional 2D.** Photographs of (**A**) out-of-plane (or multi-plane) tape printing (see movie S2); (**B**) printing on a vertical surface (see movie S2); (**C**) conformal printing on a curved surface, hemisphere (see movie S3); and (**D**) 3D printed: (i) single line (inset is the side view); (ii) overhanging structure; (iii) microscope image of the cross-section of (i); and (iv) woodpile structure. Scale bars, 3 cm (**B**, **C**, and **D** (i)), 1 cm (**D**(ii)), 1 mm (**D** (iii)), and 1 cm **D** (iv).

**Demonstration of practical applications**
At last, we explore the application potential of the developed tape placement technique through various proof-of-concept demonstrations. As shown in Fig. 5 (A and B), we successfully printed Cu tape in-plane and out-of-plane (i.e., surfaces with different heights) as conductors for a light-emitting diode (LED) circuit (movie S4) and a DC motor circuit (movie S5). Here, for out-of-plane printing, the placement starts at the bottom surface (about 4 cm) to provide sufficient adhesion force to support the pulling tension; then TPM moves to the next point at an angle to create an overhanging tape laying until it reaches the next contact point, namely, the cube surface and finally completes the placement on the cube. Then, we printed a 6×6 capacitive sensor array using two different tapes: Cu tape as conductors and vinyl tape as dielectric materials. The design schematic and fabrication of the capacitive sensor are illustrated in Fig. 5 (C). First, 6 strips of Cu tape are printed on the substrate vertically as the conductive electrode (first layer); then, 6 strips of vinyl tape are placed orthogonally on the Cu tapes as the second layer; and lastly, another layer of 6 strips of Cu tape is placed on the top of, dielectric vinyl tape. After this, the sensor array is connected to the circuit board and covered with a transparent plastic film for characterization and human-robot interaction (Fig. 5C), where the finger-touched point has a big capacitance change, showing a darker color on the computer screen. Namely, the sensor array can locate and quantify the force of the contact. To better quantify the relationship between weight (or force) and the change of capacitance, different weights are placed on top of the sensor array. Figure 5D clearly shows that the capacitance change increases with the weight (or force) and gradually saturates around 500 g (or 4.9 N). It suggests such a sensor array can also be used for force sensing in other applications. Furthermore, we integrate this sensor array with a simple robotic hand to work as a controller board (Fig. 5E, movie S6). It can be seen that different locations are codified to control different fingers (Fig. 5E (i-v)), and different forces can control the bending angle of the finger (Fig. 5E (vi-viii)). Yet, due to the interference of capacitive sensors, imperfect alignment, and relatively big space of the tapes, the performance of the sensor array can be further improved. We expect such a tape placement technique can be used to fabricate large-scale sensor-enabled interactive floors for intelligent home and immersion entertainment environments (*49–53*).

**Fig. 5. Tape printing enabled applications.** (**A**) Cu tape for high current motor wiring circuit; (**B**) Cu tape as out-of-plane conductors for LED light; (**C**) Multiple tape printing for capacitive sensor array: i. schematic drawing of sensing mechanism; ii-iv Photos of different tapes during printing, and after wiring; v sensor array application for human-robot interaction. (**D**) The sensor responds to different weights. (**E**) Sensor array used for robotic hand control. i-v control of different fingers; vi-viii control of different bending degrees of a representative finger.

## DISCUSSION

Different from automated tape placement, which mainly uses compaction force, we manipulate the tension ($F_t$) formed by the TPM to realize out-of-plane placement. To ensure full manipulation of the tension, two conditions (fig. S9) should be met, namely, (a) $F_t cos\alpha \leq \mu F_{adhesion}$ (here, $\alpha$ is the angle between tension and adhesion plane, $\mu$ is the effective adhesion coefficient associated with the vertical adhesion force $F_{adhesion}$); (b) $F_t sin\alpha \leq F_{adhesion}$. Such conditions can be met relatively easily as long as the tape feeding is smooth. In our current experiments, the tension is maintained small enough by controlling the tape feeding rate and the pulling speed of the TPM. This force analysis can be further detailed in the future for more accurate manipulation (*54, 55*). However, the current tension control is sufficient for overhanging structure fabrication, which can be easily extended to other material systems for 3D structuring. In the future, contacts should be better utilized for manufacturing through a deformable continuous object manipulation strategy (*32, 56*).

In this work, copper tape is used as a representative tape material for manipulation and manufacturing. Solid copper tape has certain advantages for conductive circuit fabrication compared to conductive inks and conductive filaments. For example, copper has low resistivity (1.68 μΩ·cm), and good flexibility, which is excellent for flexible electronics (*24*). As demonstrated in Fig. 5B, the continuous copper tape allows us to print with bending edges and overhanging structures, which is impossible for conductive inks or pastes. Due to the adhesive material, the copper tape can be readily printed on different surfaces for circuit fabrication. Besides, copper tape provides good potential for integrating electrical components due to its strong electrical and mechanical bonding resulting from good solderability (*45*). Yet, the system can be readily generalized to other tapes for broader industrial applications, such as assembly and packaging.

Although promising, we do observe some modules that require further improvement and investigation. For example, the current printer is not able to print perfect conformal traces due to the limitation of roller size and the module's volume envelope, especially at the sharp corners. This can be improved by using smaller rollers and streamlining the module volume envelope. The current printing resolution is limited by the copper tape width, which can be improved using custom tape with a smaller width. The printing accuracy can be improved if the TPM is mounted to a gantry stage and uses a better cutting mechanism, such as a laser beam (*57*).

In summary, we introduced a cost-effective printhead (i.e., TPM) mounted with a multi-axis robot system for additive manufacturing of tape materials, specifically showcasing the applications with copper tape. Our system demonstrated various print patterns on different substrates, exhibiting excellent conductivity. Leveraging the tension between the printed tape and the tape print module, we achieved out-of-plane printing without needing support, enabling overhanging structure printing. Our printing strategy offers a new paradigm for conductor manufacturing, which is easily integrable with commercial desktop 3D printers and industrial robots with reasonable customization. Additionally, this system extends to real 3D printing with tape materials. It can also be applied to other tape materials such as polyimide (PI) tape (*58*) and carbon fiber prepreg tape (*3*). We believe this innovative printing approach can broaden the application of tape materials in 3D printing and inspire new automated manufacturing strategies using dexterous robots and manipulators (*36*).

## MATERIALS AND METHODS

### Materials

The tape materials used in this work include commercially available copper tape, polyimide (PI) tape, and vinyl tape. Copper tapes with varying widths—0.25 inch (Obaka Copper Foil Tape), 0.20 inch (EDSRDRUS Copper Foil Tape), and 0.125 inch (EDSRDRUS Copper Foil Tape)—were purchased from Amazon. These copper tapes all have a thickness of approximately 50 μm and feature conductive adhesive sandwiched between the copper tape and the backing paper. We also used polyimide (PI) tape (ProTapes Pro 950 Polyimide Film Tape, 7500V Dielectric Strength) and vinyl tape (VViViD Black Carbon Fiber Air-Release Adhesive Vinyl Tape Roll) with a width of 0.25 inch, sourced from Amazon. The print module fabrication involves using an acrylic sheet with a 0.25-inch thickness (Astrariglas cast acrylic clear plexiglass) as the primary support for other components. The tool mount connecting the robot arm and the support plate is 3D printed using photopolymer resin (Clear v4, Formlabs). Various parts, such as the tape guides, holder, cutting rack-pinion, and blade guide, are 3D printed using polylactic acid (PLA, Hatchbox 1.75mm PLA 3D Printer Filament). Silicone (Elastosil M4601) is utilized to create soft contact surfaces on the rollers for tape handling and compaction (details can be found in, note S1 and fig S1).

### Tape print module

The tape print module (TPM, Fig. 1B) was fabricated in-house with various components manufactured by 3D printing, and laser cutting. An acrylic sheet (0.25 inch) was used as a backplate where the rest of the components were mounted to complete the TPM. It was attached to a 3D-printed mount connected to the robotic arm as an end-effector. Components like guiding channels, tape holder, and cutting rack-pinion housing were printed with FFF printing. Silicone polymer was used for tape-feeding contacts through mold casting. Standard nuts (M3-M5), screws, and washers were used for assemblies. Off-the-shelf components, including ball bearings (608RS 8×22×7 mm), stepper motor (Stepperonline Pancake Nema 17 Stepper Motor), servo motor (Treedix MG996R metal gear servo motor), cutting blade (X-Acto H0859 #18, heavyweight chiseling blade), and idler pulley (3Dman GT2 20 Toothless Bore 5mm), were employed for tape feeding and cutting mechanisms. The TPM weighs less than 500 g (~470 g), and the fabrication cost is around $100. More details can be found in Note S1 and Fig. S1.

### Print control

Print toolpath is generated using robot programming software and executed in real time on our system. The TPM undergoes a calibration routine to set its relative z-height to 0 with respect to the print substrate. The print process begins as the robot follows the defined toolpath and the robot controller communicates with PCM over I/O ports to regulate tape dispensing and cutting on demand. A critical challenge in ensuring dimensional accuracy and straightness is maintaining proper synchronization between the robot controller and PCM. Typically, synchronization is less problematic in systems with centralized control architectures where a single controller manages all active components [Badarinath. R, 2021]. However, in our system, control is distributed between the robot controller and PCM. The tape feeding (controlled with a stepper motor) and cutting (operated by a servo motor) processes are electronically controlled by the print module.

The detailed printing parameters are described as follows. In Fig. 2, tapes in 2A, 2C(i-ii), and 2D were printed at a speed of 25 mm/s, while 2C (iii-vi) and 2B were printed at 10 mm/s without active compaction force control. In Fig. 4, tape placements in 4A-C were printed at a speed of 25 mm/s, while 4D was printed at a speed of 10 mm/s. Active compaction control was applied in 4C and 4D, with a 4 N force (normal to the substrate), while no compaction control was employed for the rest. In Fig. 5, tape placements in 5A-E were printed at a speed of 25 mm/s without active compaction force control.

### Adhesion and roughness measurement

To study the influence of different substrates on the printed copper tape's surface roughness, we captured corresponding images and measured them with ImageJ software. Five samples of 15 cm length copper tape were printed on all three substrates under controlled photographic conditions. ImageJ software was used to segment the printed tape from the substrate, and a relative surface roughness measurement was recorded for comparison, as shown in Fig. S7. We also investigated the adhesion between the substrates and the tape through a 180° peel test. Five samples were printed on each substrate, and a mechanical tester (Mark 10 ESM303) along with a force gauge (Mark 10 Force Gauge Model M5 300) was employed to perform the 180° peel test at a peeling speed of 5 mm/s. Adhesion data were collected from the force gauge as 5 cm of printed tape was peeled off from different substrates (Fig. 3F.)

**Supplementary Materials**
- Note. S1 Module design and fabrication
- Note. S2 Tape printing quality evaluation
- Note. S3 Cartesian coordinates vs. compaction-based toolpath
- Fig. S1 Detailed design and components of the tape printing module.
- Fig. S2 TPM module control and I/O communication with robot controller
- Fig. S3 Limitations during curved printing.
- Fig. S4 Concentric circles printing.
- Fig. S5 Illustration for length and straightness measurement.
- Fig. S6 Cartesian coordinate vs. relative coordinate with compaction.
- Fig. S7 Influence of compact force on tape placement.
- Fig. S8 Procedure for surface roughness calculation.
- Fig. S9 Illustration of force analysis.
- Movie S1 Nonlinear printing
- Movie S2 Printing on vertical plane
- Movie S3 Conformal printing on hemisphere
- Movie S4 LED circuit printing
- Movie S5 Motor circuit printing
- Movie S6 Human-robot interaction with printed sensor array

   *Systems (IROS)* (IEEE, 2018; https://ieeexplore.ieee.org/document/8593780/), pp. 479–484.
30. H. Zhang, J. Ichnowski, D. Seita, J. Wang, H. Huang, K. Goldberg, "Robots of the Lost Arc: Self-Supervised Learning to Dynamically Manipulate Fixed-Endpoint Cables" in *2021 IEEE International Conference on Robotics and Automation (ICRA)* (IEEE, 2021; https://ieeexplore.ieee.org/document/9561630/), pp. 4560–4567.
31. Y. She, S. Wang, S. Dong, N. Sunil, A. Rodriguez, E. Adelson, Cable manipulation with a tactile-reactive gripper. *Int J Rob Res* **40**, 1385–1401 (2021).
32. J. Zhu, B. Navarro, R. Passama, P. Fraisse, A. Crosnier, A. Cherubini, Robotic Manipulation Planning for Shaping Deformable Linear Objects WithEnvironmental Contacts. *IEEE Robot Autom Lett* **5**, 16–23 (2020).
33. M. Kayser, L. Cai, S. Falcone, C. Bader, N. Inglessis, B. Darweesh, N. Oxman, FIBERBOTS: an autonomous swarm-based robotic system for digital fabrication of fiber-based composites. *Construction Robotics* **2**, 67–79 (2018).
34. F. Augugliaro, A. Mirjan, F. Gramazio, M. Kohler, R. D'Andrea, "Building tensile structures with flying machines" in *2013 IEEE/RSJ International Conference on Intelligent Robots and Systems* (IEEE, 2013; http://ieeexplore.ieee.org/document/6696853/), pp. 3487–3492.
35. X. Jiang, Y. Nagaoka, K. Ishii, S. Abiko, T. Tsujita, M. Uchiyama, Robotized recognition of a wire harness utilizing tracing operation. *Robot Comput Integr Manuf* **34**, 52–61 (2015).
36. S. Yuan, A. D. Epps, J. B. Nowak, J. K. Salisbury, "Design of a Roller-Based Dexterous Hand for Object Grasping and Within-Hand Manipulation" in *2020 IEEE International Conference on Robotics and Automation (ICRA)* (IEEE, 2020; https://ieeexplore.ieee.org/document/9197146/), pp. 8870–8876.
37. T. Ozaki, T. Suzuki, T. Furuhashi, S. Okuma, Y. Uchikawa, Trajectory control of robotic manipulators using neural networks. *IEEE Transactions on Industrial Electronics* **38**, 195–202 (1991).
38. A. H. Khan, S. Li, X. Luo, Obstacle Avoidance and Tracking Control of Redundant Robotic Manipulator: An RNN-Based Metaheuristic Approach. *IEEE Trans Industr Inform* **16**, 4670–4680 (2020).
39. R. Badarinath, V. Prabhu, Integration and evaluation of robotic fused filament fabrication system. *Addit Manuf* **41**, 101951 (2021).
40. P. Urhal, A. Weightman, C. Diver, P. Bartolo, Robot assisted additive manufacturing: A review. *Robot Comput Integr Manuf* **59**, 335–345 (2019).
41. M. Safeea, R. Bearee, P. Neto, An integrated framework for collaborative robot-assisted additive manufacturing. *J Manuf Process* **81**, 406–413 (2022).
42. M. Doshi, A. Mahale, S. Kumar Singh, S. Deshmukh, Printing parameters and materials affecting mechanical properties of FDM-3D printed Parts: Perspective and prospects. *Mater Today Proc* **50**, 2269–2275 (2022).
43. S. G. Lee, S. W. Lye, Design for manual packaging. *International Journal of Physical Distribution & Logistics Management* **33**, 163–189 (2003).
44. V. Savage, R. Schmidt, T. Grossman, G. Fitzmaurice, B. Hartmann, "A series of tubes" in *Proceedings of the 27th Annual ACM Symposium on User Interface Software and*

**Acknowledgments:** We would like to thank Walmart for the donation of the UR10E robotic arm. This project was supported by the Pandemic Research Recovery Grant and startup fund from the University of Arkansas. This project is also partially supported by the University of Arkansas Engineering Research Seed Grant and USDA-NIFA (award # 2023-67022-39074).

**Funding:**

National Institute of Food and Agriculture of the U.S. Department of Agriculture grant 2023-67022-39074 (WS); University of Arkansas, Pandemic Research Recovery Grant (WS); University of Arkansas, Engineering Research Seed Grant (WS).

**Author contributions:**
Conceptualization: WS
Data curation: NT
Formal analysis: NT
Investigation: NT
Methodology: NT, WS
Project administration: WS
Resources: WS
Software: NT
Supervision: WS
Validation: NT, WS
Visualization: NT
Writing – original draft: NT, WS
Writing – review & editing: NT, RW, YS, WZ, WS

**Competing interests:** Authors declare that they have no competing interests.

**Data and materials availability:** All data are available in the main text or the supplementary materials. Additional data can be requested from the corresponding author.

**Figures:**

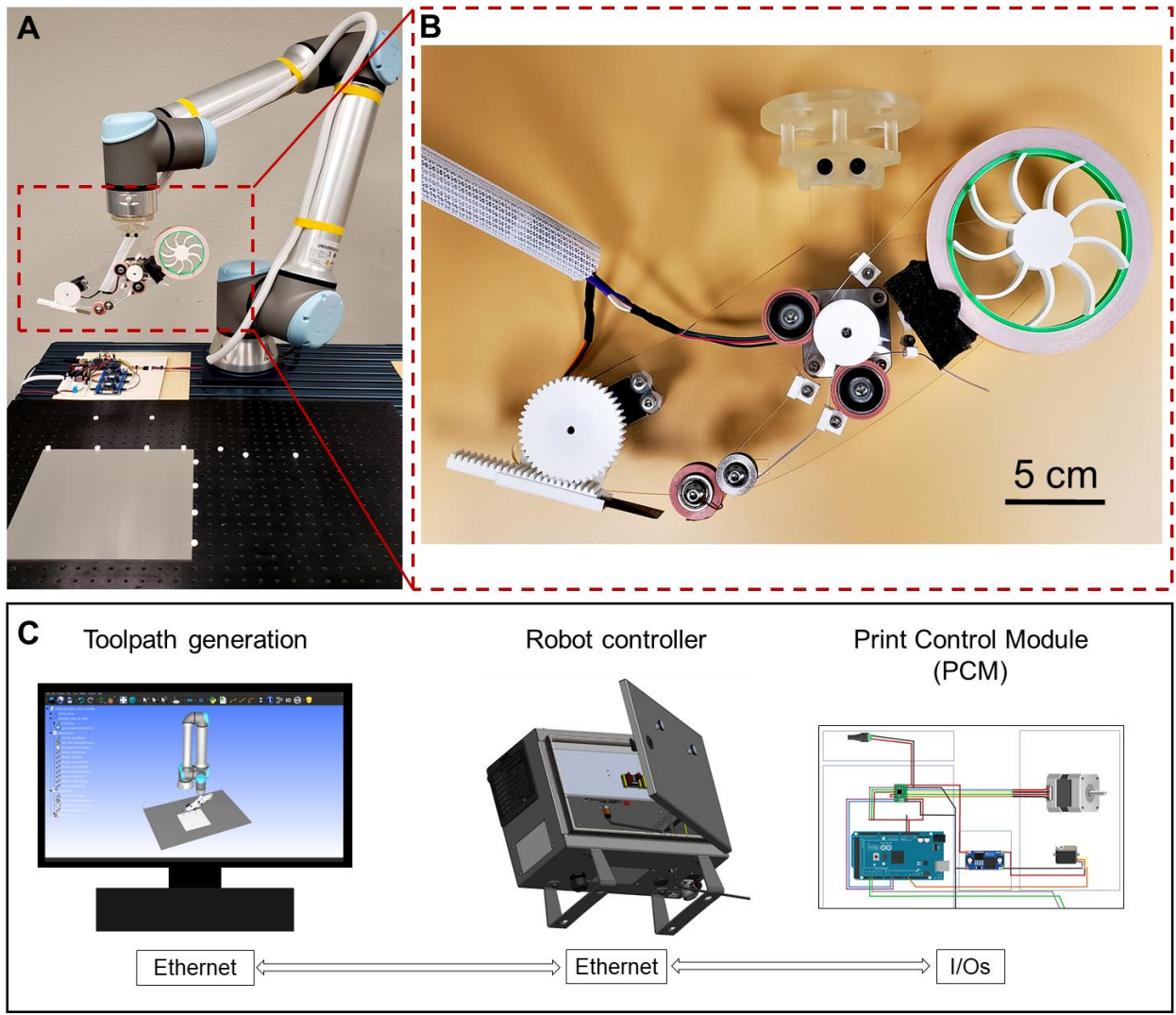

**Fig. 1. Overview of tape manipulation and 3D printing system.** (**A**) Assembled tape placement module (TPM) on a 6-axis robot. (**B**) Photograph of the TPM end-effector. (**C**) System architecture overview. (More details can be found in fig. S2).

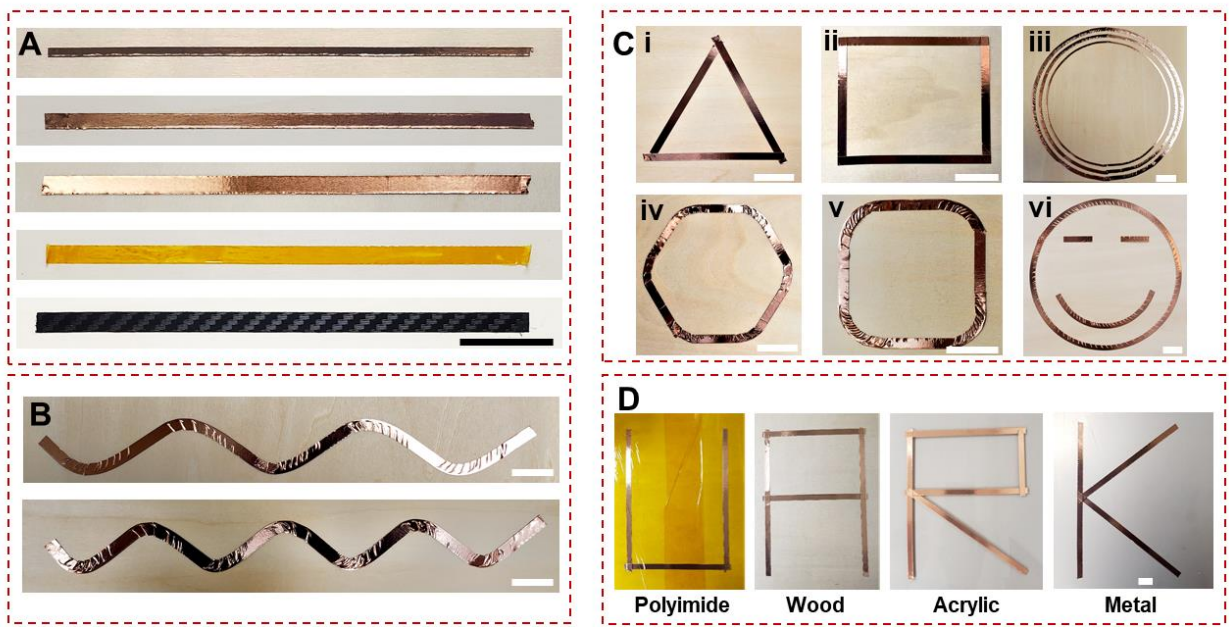

**Fig. 2. Representative Cu tape 2D printing.** (**A**) Straight lines of Cu tapes (with different widths), and other tape materials (e.g., polyimide and vinyl). (**B**) Nonlinear printing of wavy lines. (**C**) Common geometries printed with Cu tape: segment-by-segment printing of a (i) triangle and (ii) rectangle; continuous printing of (iii) circles, (iv) hexagon, and (v) rectangle; (vi) a smiley face. (**D**) Cu tape printed "UARK" on different substrates. (scale bar, 3 cm.)

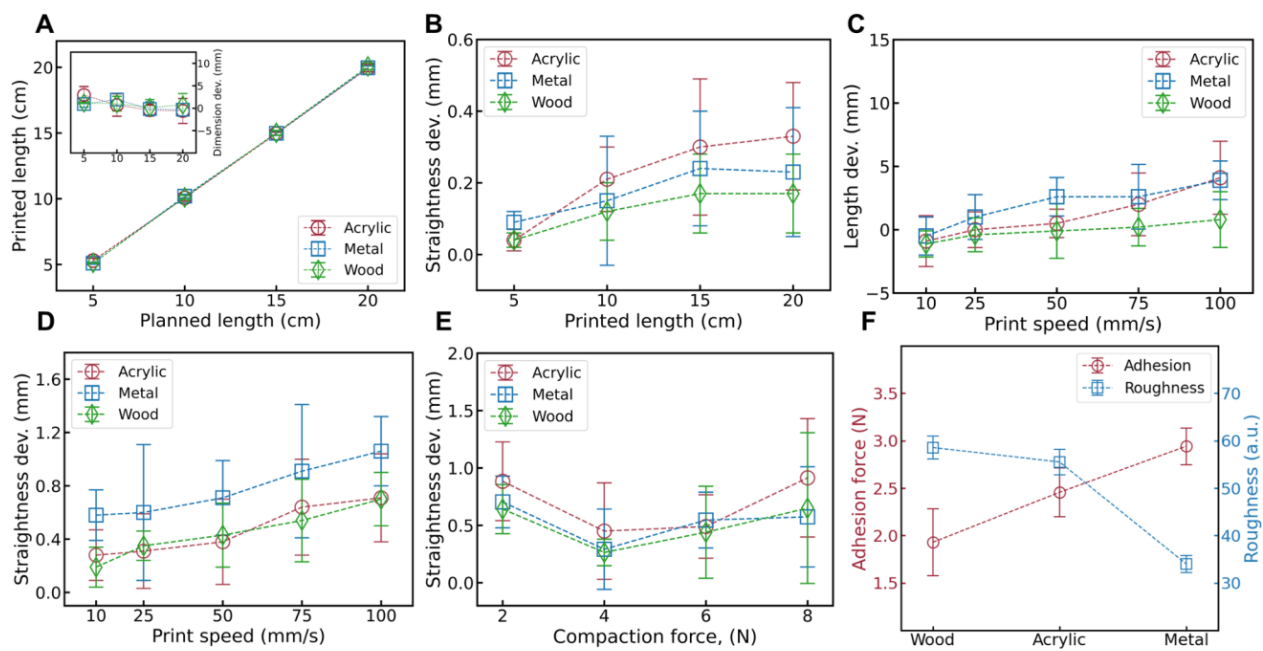

**Fig. 3. The influence of Cu tape printing parameters on quality.** (**A**) print length consistency (print speed 25 mm/s, point-to-point toolpath trajectory); (**B**) straightness consistency (print speed 25 mm/s, point-to-point toolpath trajectory); (**C**) print length consistency and (**D**) straightness consistency with different print speeds with a fixed length of 15 cm (point-to-point toolpath trajectory); (**E**) print straightness consistency with variable compaction force on the substrate (relative point to point toolpath trajectory with compaction force applied normal to the substrate); (**F**) tape adhesion and roughness on different substrates (details of measurement can be found in Materials and Methods section and fig S8).

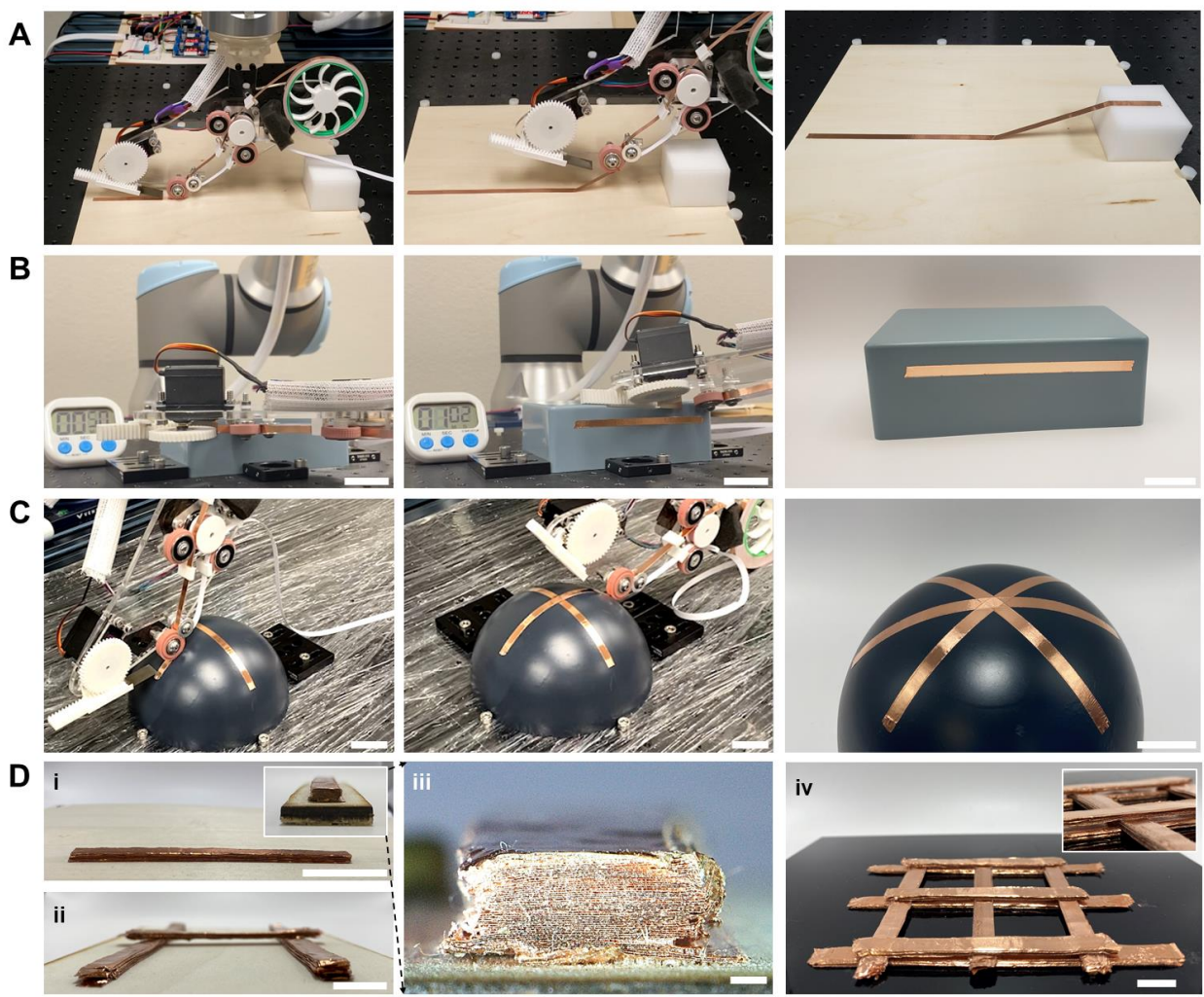

**Fig. 4. Printing beyond traditional 2D.** Photographs of (**A**) out-of-plane (or multi-plane) tape printing (see movie S2); (**B**) printing on a vertical surface (see movie S2); (**C**) conformal printing on a curved surface, hemisphere (see movie S3); and (**D**) 3D printed: (i) single line (inset is the side view); (ii) overhanging structure; (iii) microscope image of the cross-section of (i); and (iv) woodpile structure. Scale bars, 3 cm (**B**, **C**, and **D** (i)), 1 cm (**D**(ii)), 1 mm (**D** (iii)), and 1 cm **D** (iv).

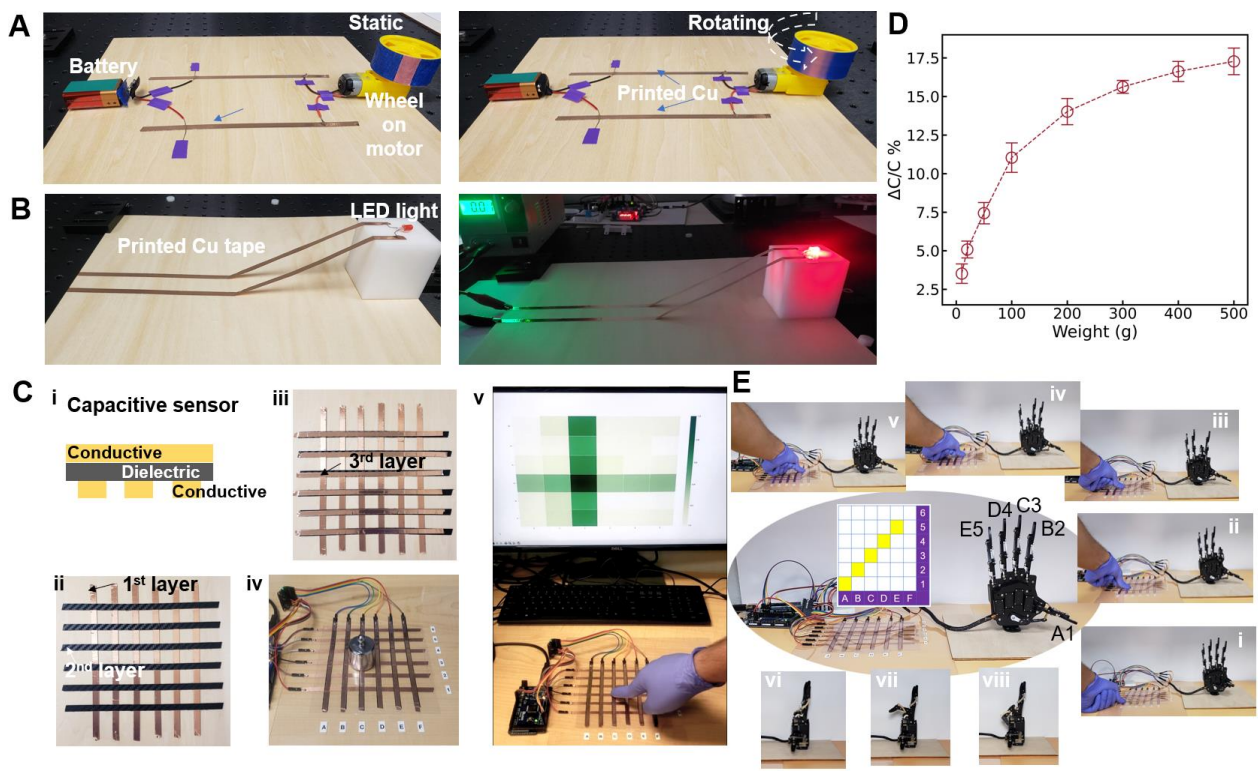

**Fig. 5. Tape printing enabled applications.** (**A**) Cu tape for high current motor wiring circuit; (**B**) Cu tape as out-of-plane conductors for LED light; (**C**) Multiple tape printing for capacitive sensor array: i. schematic drawing of sensing mechanism; ii-iv Photos of different tapes during printing, and after wiring; v sensor array application for human-robot interaction. (**D**) The sensor responds to different weights. (**E**) Sensor array used for robotic hand control. i-v control of different fingers; vi-viii control of different bending degrees of a representative finger.

Supplementary Materials for

# Robot Tape Manipulation for 3D Printing

Nahid Tushar, Rencheng Wu, Yu She, Wenchao Zhou, Wan Shou*

Corresponding author: Wan Shou, wshou@uark.edu

**The PDF file includes:**

**Other Supplementary Material for this manuscript includes the following:**

## Note S1 Module design and fabrication

Fig. S1 displays the prototype CAD design, while Fig. 1 shows the fabricated module. Tape feeding is governed by two key factors: tension and feeding force. Tension is generated on the copper tape between the printed trace and the printhead through the compaction roller, allowing the creation of lines between two points, regardless of the structure in between. The feeding force is supplied by a pair of rollers, driven by a stepper motor (as shown in fig. S1), guiding the tape into the print path. With these capabilities, our printhead can easily integrate with a robotic arm or desktop-scale 3D printers to apply conductors to desired surfaces or points, serving various electronic component fabrication and integration purposes. Tension is generated at specific points, such as the compaction roller and the feeding wheel for the copper tape, the separator guide, and the support roller for the back paper. Primary control of feeding is handled by the main roller (as depicted in fig. S1). The compaction roller, also shown in fig. S1, enhances tape adhesion on the intended substrate for secure placement. The main rollers, support rollers, and compaction rollers feature silicone-based surfaces to ensure proper contact, feeding, and tension. The cutting module, as depicted in fig. S1, utilizes a rack and pinion setup with an x-acto blade, powered by a servo motor, precisely positioned to cut and secure placed tape on the substrate. In summary, the printhead is equipped with a tape holder for installing copper tape (or other tapes), four tape guides, a back paper pulley/guide, a stepper motor-controlled feeding wheel with threaded silicone as a contact surface, and a silicone-based compaction roller with a wire guide for tape. Additionally, a servo motor-driven, pinion-based cutting assembly is included within the printhead.

## Note S2 Tape printing quality evaluation

To assess the system's performance, we conducted a comprehensive evaluation comparing planned and printed tape deviations in terms of length and alignment, using copper tape as the test material. Copper tape was printed in various lengths (5, 10, 15, and 20 cm) on different substrates (metal, acrylic, and wood). For each length, nine samples were printed on each substrate, and their values were recorded. To account for uneven endpoints, we measured the effective length and maximum alignment deviation, as shown in Fig. S4. The mean and sample standard deviation data for all prints and substrates are illustrated in the system study (Fig. 3A-B). We also examined the effect of different print speeds, ranging from 10 to 100 mm/s, for five 15 cm long copper tape samples printed on all substrates (acrylic, metal, wood). The results are presented in Fig. 3C-D. In addition to printing along toolpath-based Cartesian coordinates, we experimented with printing using relative coordinates and varying compaction force (normal force exerted on the substrate by the TPM) at levels of 2, 4, 6, and 8 newtons (N). Five samples of 15 cm length copper tape were printed on all three substrates, and their length and straightness accuracies were recorded and are displayed in Fig. 3E and Fig. S6.

## Note S3 Cartesian coordinates vs. compaction-based toolpath

In our system, tape print toolpaths can be realized using two methods: (1) Cartesian coordinates for point-to-point printing, or (2) relative start and end points, with a compaction force applied normally to the substrate using the robot's force-torque sensor (UR10E's force torque sensor). A study comparing the effects of both methods is presented in fig. S6. Copper tape, ranging from 5 to 20 cm in length, was printed on acrylic, metal, and wood substrates (nine prints for each length). Measurements of length, straightness deviation, and surface roughness of the placed tape were conducted and compared (see Fig. S6). Our results did not show a significant difference between Cartesian coordinate-based and compaction force-based printing. However, compaction-based printing offers advantages for specialty applications, such as 3D printing of copper tape through layer-by-layer placement and printing on curved surfaces, such as hemispheres, as demonstrated in Fig 4D.

**Supplementary Figures and Tables**

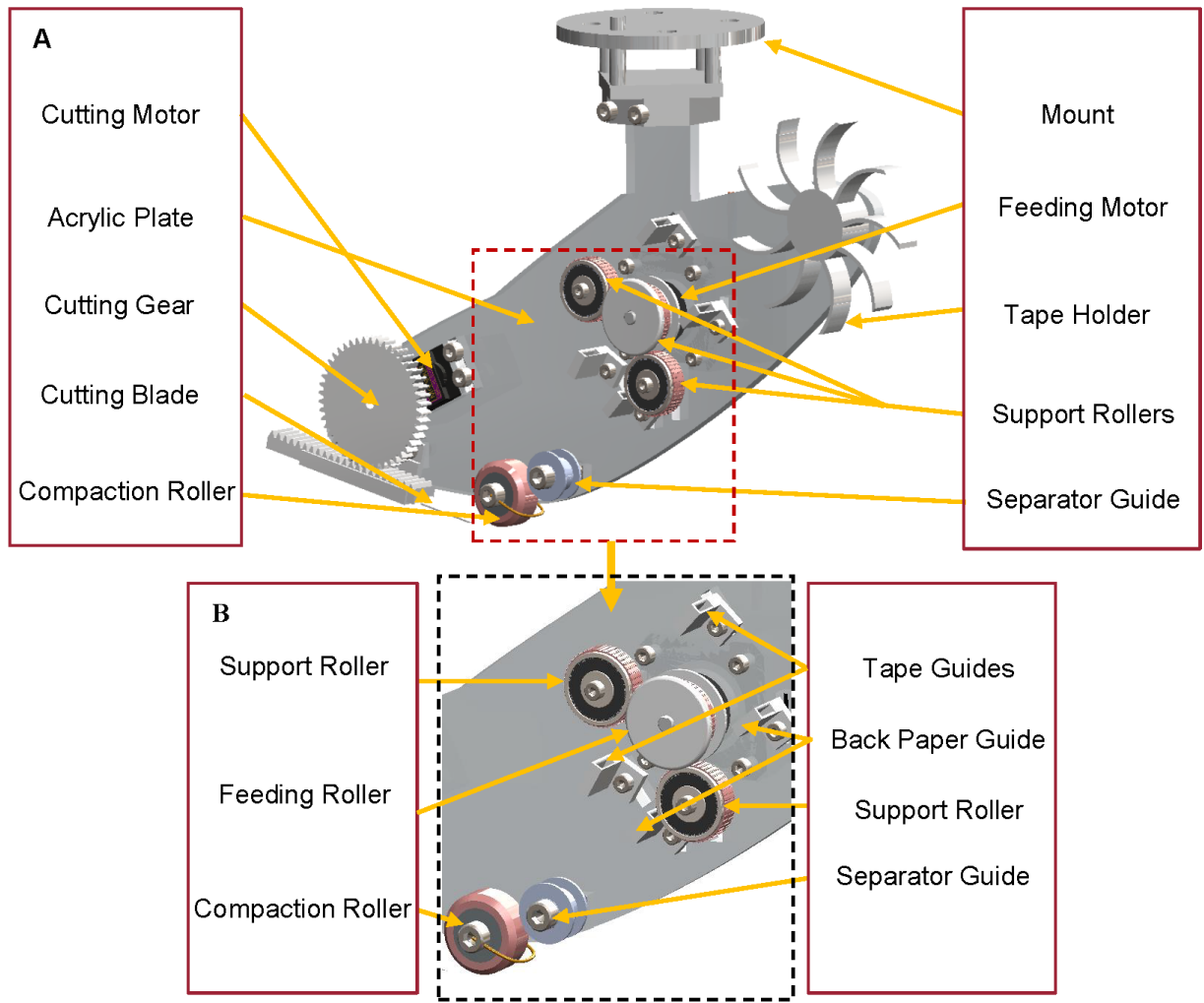

**Fig. S1 Detailed design and components of the tape printing module.**

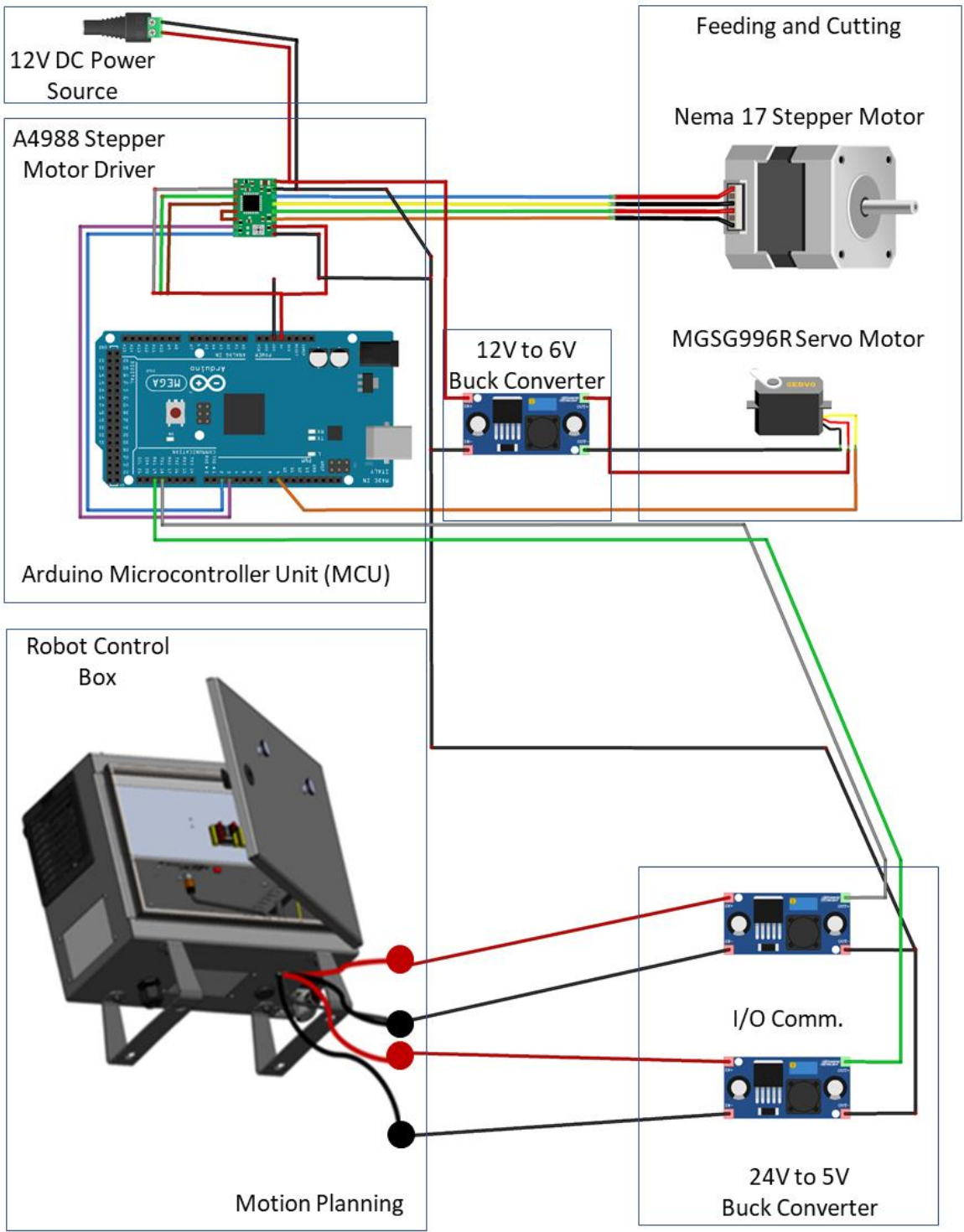

**Fig. S2 TPM control and I/O communication with robot controller.**

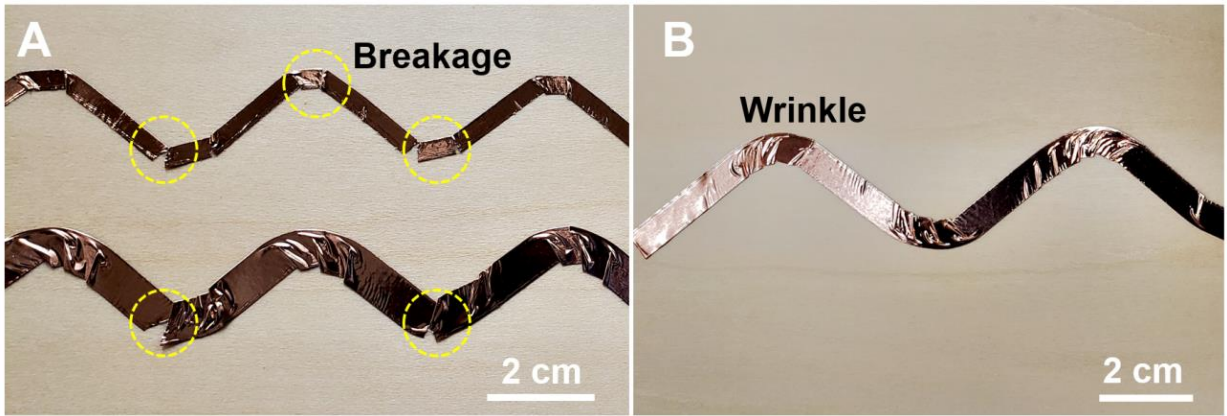

**Fig. S3 Limitations during curved printing.** (A) Breakage of tape with too small curvature. (B) Wrinkles and warps at turns.

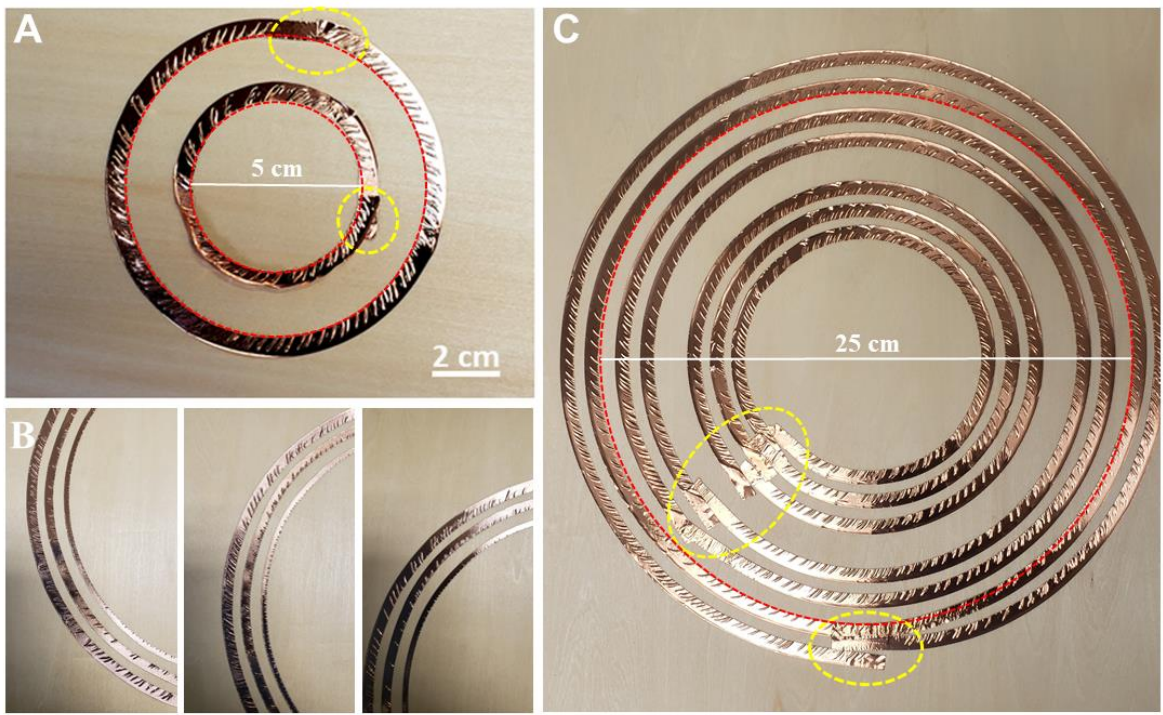

**Fig. S4 Concentric circles printing.** **(A)** Circle toolpath printing with 10 cm and 5 cm diameters (starting and ending points are shown in yellow dashed circles, representing the misalignment issue due to robot joint limit). **(B)** Demonstration of circular pattern with reasonably uniform tape placement. **(C)** Repetitive concentric tape placement. (dashed yellow circles indicate the alignment issue at the start and end boundary)

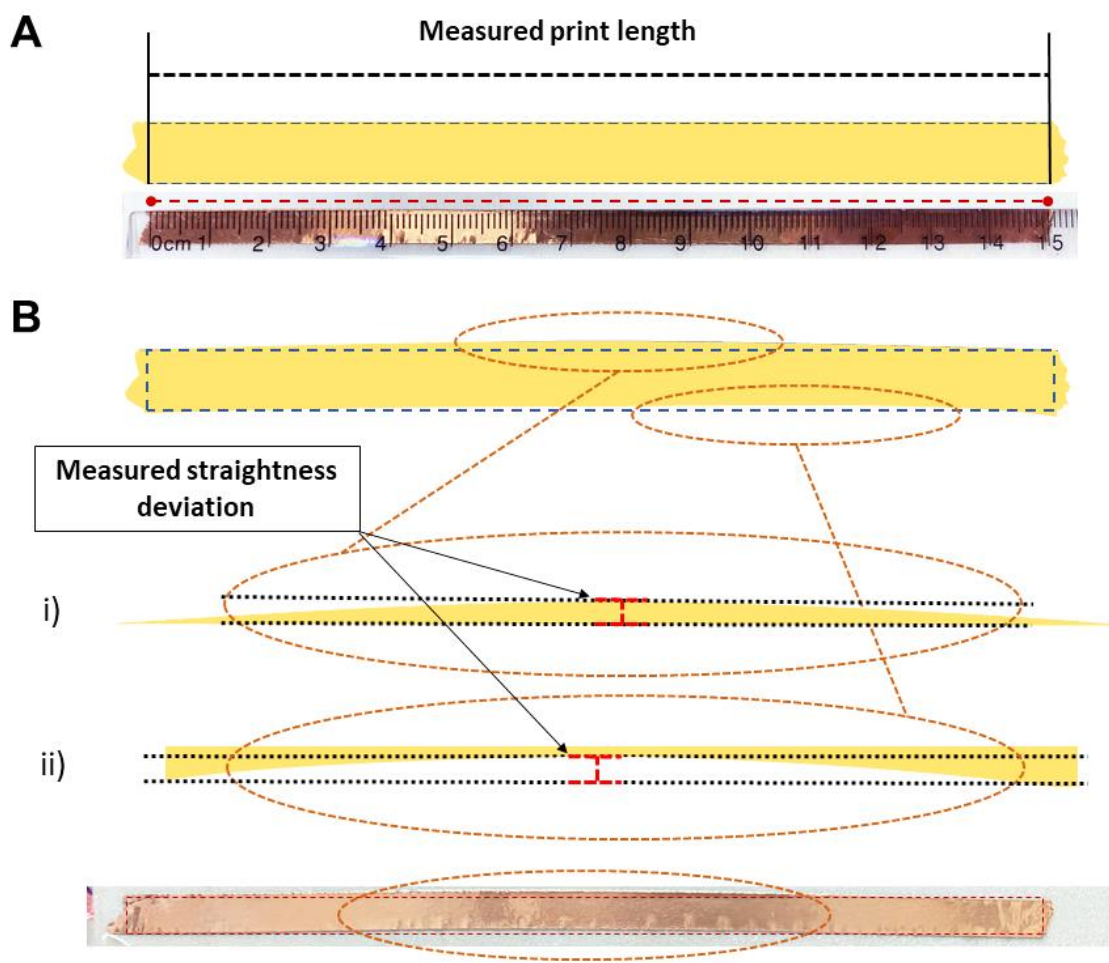

**Fig. S5 Illustration for length and straightness measurement.** (A) Effective length measurement process. (B) Straightness measurements process: Two curved surfaces (i and ii) are measured and compared. The larger measurement is recorded as the straightness deviation value.

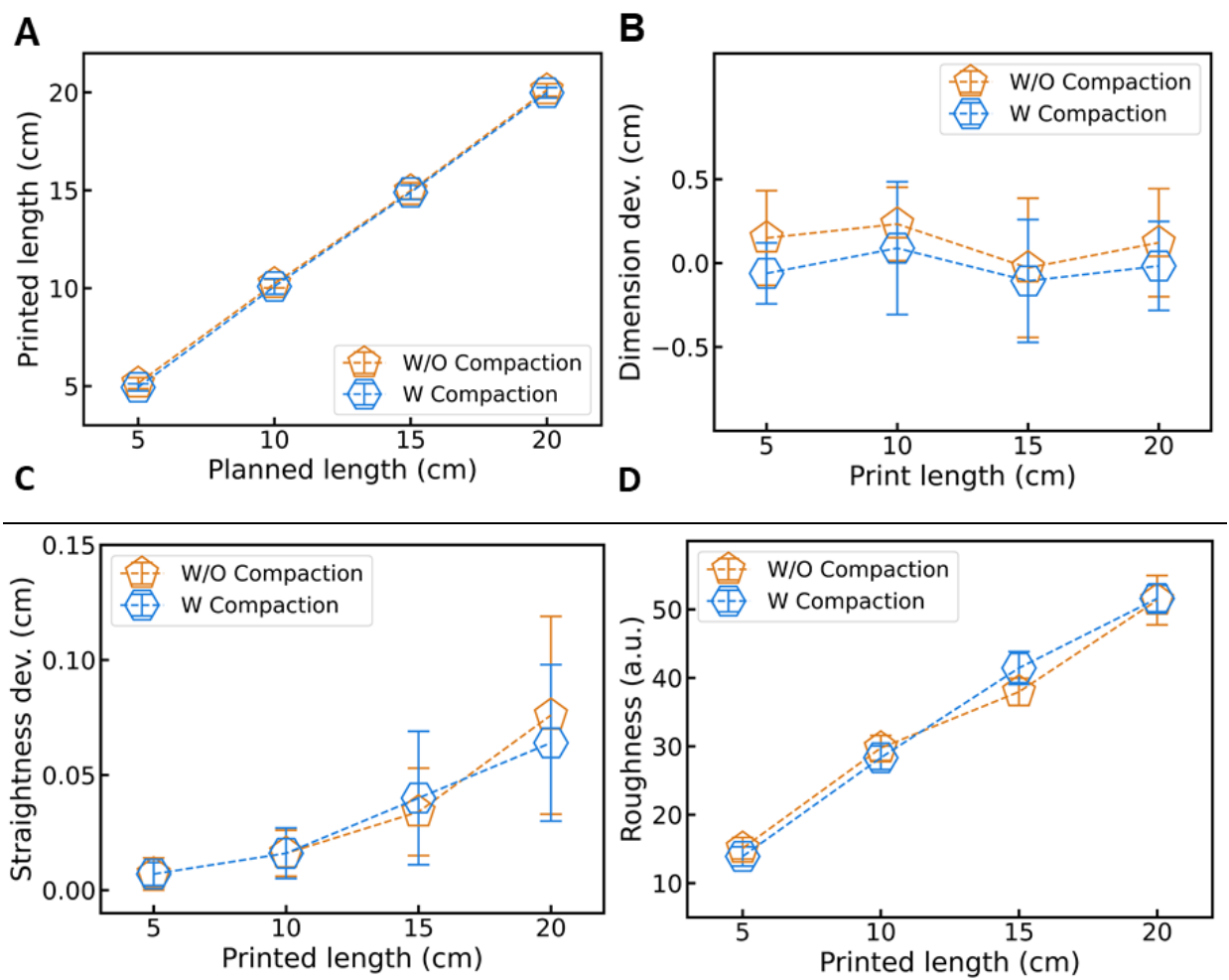

**Fig. S6 Comparison between Cartesian coordinate printing and relative coordinate printing with compaction.** **(A)**-**(B)** Print length consistency comparison; **(C)** straightness consistency comparison; **(D)** roughness comparison. ("W/O Compaction" means Cartesian coordinate printing; "W Compaction" means relative coordinate printing with compaction force applied normally to the substrate.)

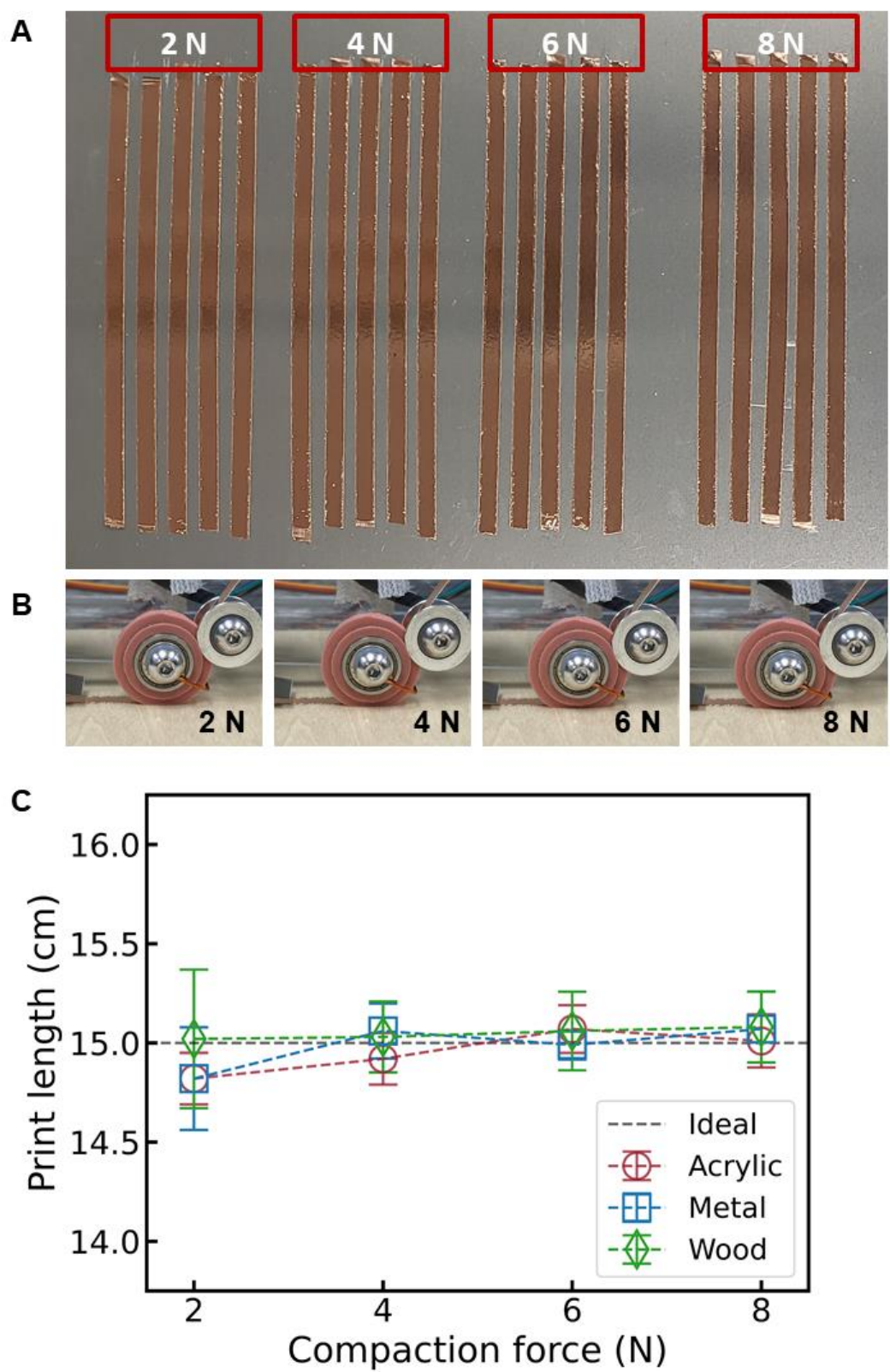

**Fig. S7 Influence of compaction force on tape placement.** (A) Photos of representative printed Cu tape (15 cm length) with varying compaction force from 2 to 8 N; (B) the shape of compaction roller under different compaction forces; (C) Influence of compaction force on printed length with a target print length of 15 cm.

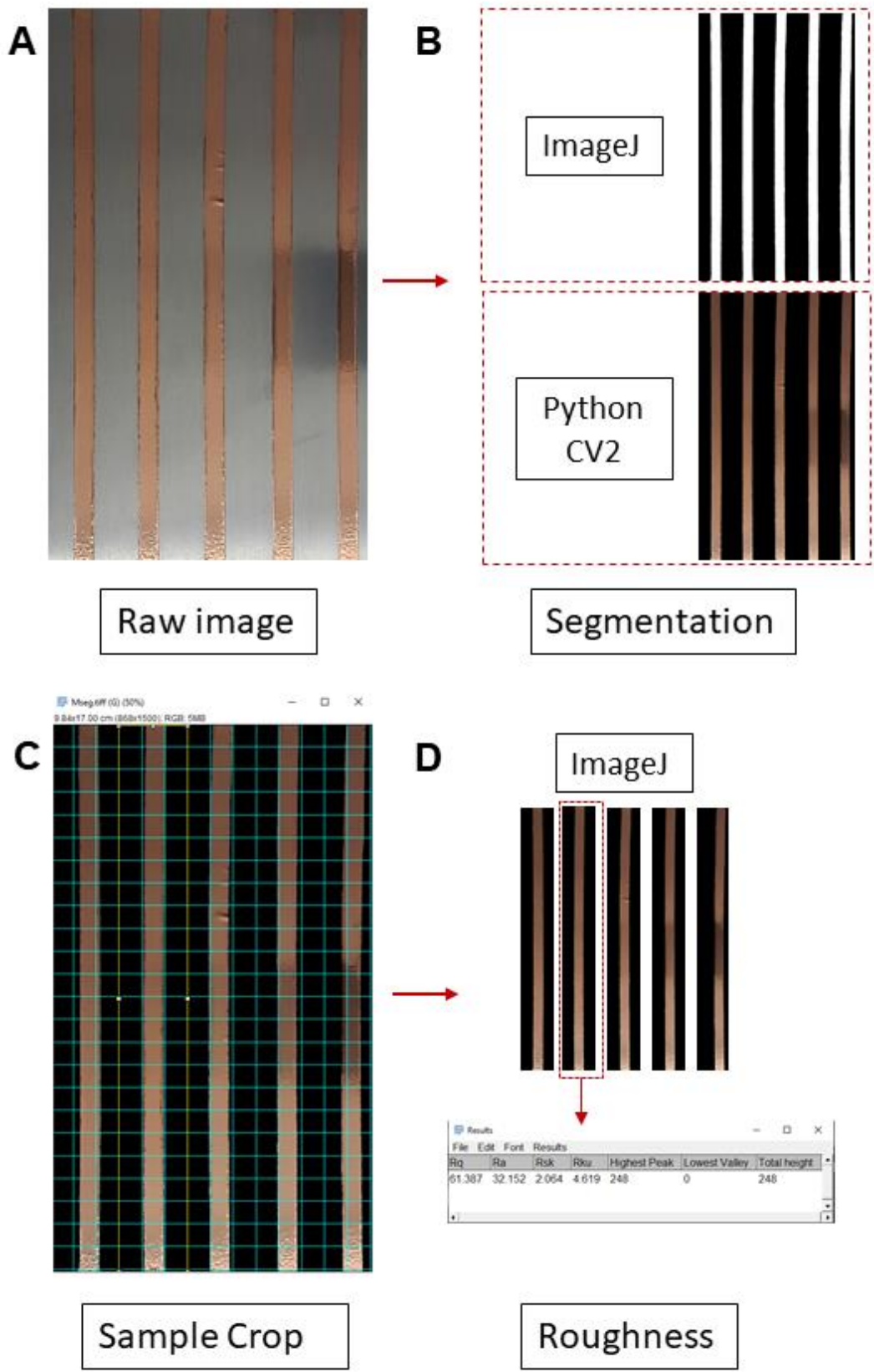

**Fig. S8 Procedure for surface roughness calculation. (A)** Raw image capture. **(B)** Image masking and background segmentation. **(C)** Sample segmentation. **(D)** ImageJ roughness measurement.

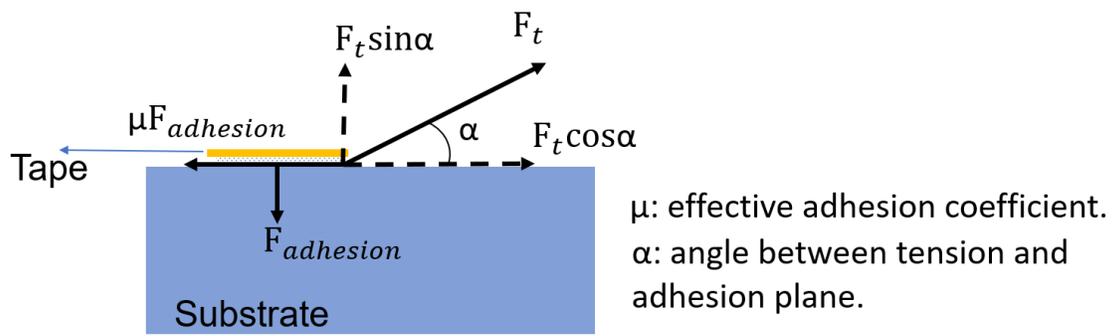

**Fig. S9 Illustration of force analysis during tape placement.**

**Legends for movies S1 to S5;**

**Movie S1. Nonlinear Printing.**
Description: Tape printing of wavy lines with varying amplitudes and wavelengths following a nonlinear toolpath.

**Movie S2. Printing on a Vertical Plane.**
Description: Tape printing as a conductive circuit trace on a vertical surface.

**Movie S3. Conformal Printing on Hemisphere.**
Description: Conformal tape placement achieved on a curved hemisphere surface.

**Movie S4. Circuit Printing for LED**
Description: Copper tape printing for an LED circuit.

**Movie S5. Circuit Printing for Motor.**
Description: Copper tape printing for a DC motor circuit.

**Movie S6. Human-Robot Interaction with Printed Sensor Array.**
Description: Sensor array used for human-robot interaction. Control of different fingers and bending degrees.